\documentclass[10pt,twocolumn,letterpaper]{article}

\usepackage[pagenumbers]{cvpr}

\usepackage{graphicx}
\usepackage{amsmath}
\usepackage{amssymb}
\usepackage{xcolor}
\usepackage{pifont}
\newcommand{\cmark}{\ding{51}}%
\newcommand{\xmark}{\ding{55}}%
\usepackage[hypcap=false]{caption}  
\usepackage{booktabs}
\graphicspath{ {./figures/} }
\usepackage{enumitem}
\usepackage[toc,page]{appendix}

\newcommand\blfootnote[1]{ 
  \begingroup
  \renewcommand\thefootnote{}\footnote{#1} %
  \addtocounter{footnote}{-1} %
  \endgroup
}

\usepackage[pagebackref,breaklinks,colorlinks]{hyperref}

\usepackage[capitalize]{cleveref}
\crefname{section}{Sec.}{Secs.}
\Crefname{section}{Section}{Sections}
\Crefname{table}{Table}{Tables}
\crefname{table}{Tab.}{Tabs.}

\begin{document}

\title{ObjectStitch: Generative Object Compositing}

\author{Yizhi Song$^{1*}$, Zhifei Zhang$^2$, Zhe Lin$^2$, Scott Cohen$^2$, Brian Price$^2$, \\
Jianming Zhang$^2$, Soo Ye Kim$^2$, Daniel Aliaga$^1$ \\
Purdue University$^1$, Adobe Research$^2$ \\
{\tt\small $^1$\{song630,aliaga\}@purdue.edu} \\
{\tt\small $^2$\{zzhang,zlin,scohen,bprice,jianmzha,sooyek\}@adobe.com}
}


\twocolumn[{
\renewcommand\twocolumn[1][]{#1}
\maketitle
\begin{center}
    \includegraphics[width=1.0\linewidth]{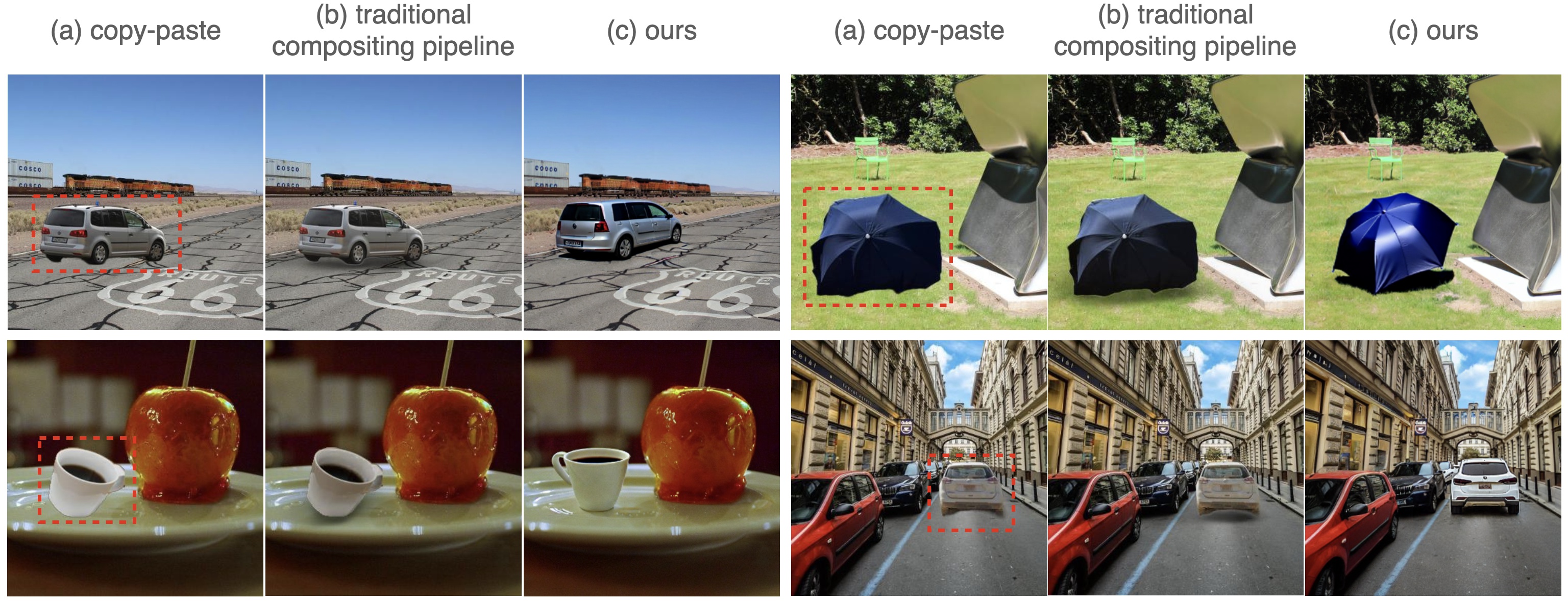}
    \captionof{figure}{Example results of object compositing with (a) copy-and-paste scheme, (b) traditional compositing pipeline and (c) ours (ObjectStitch). Traditional compositing pipeline is done with best possible off-the-shelf models including foreground/background color harmonization \cite{tsai2017deep, jiang2021ssh}, poisson blending \cite{perez2003poisson}, and shadow synthesis \cite{sheng2022controllable}. ObjectStitch achieves more realistic results, and can address geometry correction, harmonization, shadow generation, and view synthesis all-in-one while preserving similar appearance to the reference object.}
    \label{fig:teaser}
\end{center}
}]


\blfootnote{* Work done during internship at Adobe.}


\begin{abstract}
Object compositing based on 2D images is a challenging problem since it typically involves multiple processing stages such as color harmonization, geometry correction and shadow generation to generate realistic results. Furthermore, annotating training data pairs for compositing requires substantial manual effort from professionals, and is hardly scalable. Thus, with the recent advances in generative models, in this work, we propose a self-supervised framework for object compositing by leveraging the power of conditional diffusion models. Our framework can hollistically address the object compositing task in a unified model, transforming the viewpoint, geometry, color and shadow of the generated object while requiring no manual labeling. To preserve the input object's characteristics, we introduce a content adaptor that helps to maintain categorical semantics and object appearance. A data augmentation method is further adopted to improve the fidelity of the generator. Our method outperforms relevant baselines in both realism and faithfulness of the synthesized result images in a user study on various real-world images.
\end{abstract}

\section{Introduction}
\label{sec:intro}

Image compositing is an essential task in image editing that aims to insert an object from a given image into another image in a realistic way. Conventionally, many sub-tasks are involved in compositing an object to a new scene, including color harmonization \cite{jiang2021ssh, cong2020dovenet, cong2021bargainnet, xue2022dccf}, relighting \cite{zhou2019deep}, and shadow generation \cite{sheng2021ssn, hong2022shadow, liu2020arshadowgan} in order to naturally blend the object into the new image. As shown in \cref{tab:prev_works}, most previous methods \cite{lin2018st, jiang2021ssh, cong2020dovenet, cong2021bargainnet, sheng2021ssn, hong2022shadow} focus on a single sub-task required for image compositing. Consequently, they must be appropriately combined to obtain a composite image where the input object is re-synthesized to have the color, lighting and shadow that is consistent with the background scene. As shown in \cref{fig:teaser}, results produced in this way still look unnatural, partly due to the viewpoint of the inserted object being different from the overall background.

\begin{table}
  \small
  \centering
  \begin{tabular}{lllll}
    \toprule
    Method & Geometry & Light & Shadow & View \\
    \midrule
    ST-GAN \cite{lin2018st} & \cmark & \xmark & \xmark & \xmark \\
    SSH \cite{jiang2021ssh} & \xmark & \cmark & \xmark & \xmark \\
    DCCF \cite{xue2022dccf} & \xmark & \cmark & \xmark & \xmark \\
    SSN \cite{sheng2021ssn} & \xmark & \xmark & \cmark & \xmark \\
    SGRNet \cite{hong2022shadow} & \xmark & \xmark & \cmark & \xmark \\
    GCC-GAN \cite{chen2019toward} & \cmark & \cmark & \xmark & \xmark \\
    \midrule
    Ours & \cmark & \cmark & \cmark & \cmark \\
    \bottomrule
  \end{tabular}
  \vspace{-5pt}
  \caption{Prior works only focus on one or two aspects of object compositing, and they cannot synthesize novel views. In contrast, our model can address all perspectives as listed.}
  \label{tab:prev_works}
\end{table}

Harmonizing the geometry and synthesizing novel views have often been overlooked in 2D image compositing, which require an accurate understanding of both the geometry of the object and the background scene from 2D images. Previous works \cite{karsch2011rendering, karsch2014automatic, hold2017deep} handle 3D object compositing with explicit background information such as lighting positions and depth. More recently, \cite{lin2018st} utilize GANs to estimate the homography of the object. However, this method is limited to placing furniture in indoor scenes. In this paper, we propose a generic image compositing method that is able to harmonize the geometry of the input object along with color, lighting and shadow with the background image using a diffusion-based generative model.

In recent years, generative models such as GANs \cite{brock2018large, goodfellow2020generative, isola2017image, karras2019style} and diffusion models \cite{ho2020denoising, rombach2022high, avrahami2022blended, ruiz2022dreambooth, liu2021more, nichol2021glide, meng2021sdedit} have shown great potential in synthesizing realistic images. In particular, diffusion model-based frameworks are versatile and outperform various prior methods in image editing \cite{meng2021sdedit, kawar2022imagic, avrahami2022blended} and other applications \cite{poole2022dreamfusion, luo2021diffusion}. However, most image editing diffusion models focus on using text inputs to manipulate images \cite{nichol2021glide, kawar2022imagic, feng2022ernie, balaji2022ediffi, avrahami2022blended}, which is insufficient for image compositing as verbal representations cannot fully capture the details or preserve the identity and appearance of a given object image.
There have been recent works \cite{ruiz2022dreambooth, kawar2022imagic} focusing on generating diverse contexts while preserving the key features of the object; however, these models are designed for a different task than object compositing. Furthermore, \cite{kawar2022imagic} requires fine-tuning the model for each input object and \cite{ruiz2022dreambooth} also needs to be fine-tuned on multiple images of the same object. Therefore they are limited for general object compositing.

In this work, we leverage diffusion models to simultaneously handle multiple aspects of image compositing such as color harmonization, relighting, geometry correction and shadow generation. With image guidance rather than text guidance, we aim to preserve the identity and appearance of the original object in the generated composite image.

Specifically, our model synthesizes a composite image given (i) a source object, (ii) a target background image, and (iii) a bounding box specifying the location to insert the object. The proposed framework consists of a content adaptor and a generator module: the content adaptor is designed to extract a representation from the input object containing both high-level semantics and low-level details such as color and shape; the generator module preserves the background scene while improving the generation quality and versatility. Our framework is trained in a fully self-supervised manner and no task-specific labeling is required at any point during training. Moreover, various data augmentation techniques are applied to further improve the fidelity and realism of the output.
We evaluate our proposed method on a real-world dataset closely simulating real use cases for image compositing.

Our contributions are summarized as follows:
\begin{itemize}[noitemsep, topsep=3pt]
    \item We present the first diffusion model-based framework for \textit{generative object compositing} that can handle multiple aspects of compositing such as viewpoint, geometry, lighting and shadow.
    \item We propose a content adaptor module which learns a descriptive multi-modal embedding from images, enabling image guidance for diffusion models.
    \item Our framework is trained in a self-supervised manner without any task-specific annotations, employing data augmentation techniques to improve the fidelity of generation.
    \item We collect a high-resolution real-world dataset for object compositing with diverse images, containing manually annotated object scales and locations. 
\end{itemize}

\section{Related Work}
\label{sec:related}

\subsection{Image Compositing}
Image compositing is a challenging task in reference-guided image editing, where the object in a given foreground image is to be inserted into a background image. The generated composite image is expected to look realistic, with the appearance of the original object being preserved. Prior works often focus on an individual aspect of this problem, such as geometric correction \cite{lin2018st, azadi2020compositional}, image harmonization \cite{xue2022dccf, cong2020dovenet, cong2021bargainnet}, image matting \cite{xu2017deep} and shadow generation \cite{liu2020arshadowgan}.

Lin \etal~\cite{lin2018st} propose an iterative GAN-based framework to correct geometric inconsistencies between object and background images. In their model, STNs~\cite{jaderberg2015spatial} are integrated into the generator network to predict a series of warp updates. Although their work can improve the realism of the scene geometry in the composite image, it is limited to inserting furniture into indoor scenes.
Azadi \etal~\cite{azadi2020compositional} focus on binary composition, using a composition-decomposition network to capture interactions between a pair of objects.

Image harmonization aims at minimizing the inconsistency between the input object and the background image. Traditional methods~\cite{sunkavalli2010multi, lalonde2007using} usually concentrate on obtaining color statistics and then transfer this information between foreground and background. Recent works seek to solve this problem via deep neural networks. \cite{cong2021bargainnet, cong2020dovenet} reformulate image harmonization as a domain adaptation problem, while \cite{jiang2021ssh} converts it to a representation fusing problem and utilizes self-supervised training.

Shadow synthesis is an effect that is often overlooked in previous image compositing methods although it is essential for generating realistic composite images. SGRNet \cite{hong2022shadow} divides this task into a shadow mask generation stage and a shadow filling stage. SSN \cite{sheng2021ssn}, focusing on soft shadows, predicts an ambient occlusion map as cue for shadow generation.

There are also works simultaneously addressing multiple sub-problems of image compositing. Chen \etal~\cite{chen2019toward} develop a system including multiple network modules to produce plausible images with geometric, color and boundary consistency. However, the pose of the generated object is constrained by the mask input and the model cannot generalize to non-rigid objects such as animals.
Our proposed model also handles the image composition problem in a unified manner, generating foreground objects which are harmonious and geometrically consistent with the background while synthesizing novel views and shadows.

\subsection{Guided Image Synthesis}
Recent years have witnessed great advances in diffusion models. Diffusion models are a family of deep generative models based on several predominant works \cite{sohl2015deep, song2019generative, ho2020denoising}, defined with two Markov chains. In the forward process, they gradually add noise to the data, and in the reverse process, they learn to recover the data from noise. Following the growth of research in this area, diffusion models have shown their potential in a great number of applications, such as text-to-image generation \cite{nichol2021glide, feng2022ernie, rombach2022high}, image editing \cite{avrahami2022blended, kawar2022imagic, meng2021sdedit} and guided image synthesis \cite{liu2021more}.

Text-to-image synthesis has been a popular research topic over the past few years. Stable Diffusion \cite{rombach2022high} made a significant contribution to this task and it reduces the computational cost by applying the diffusion model in the latent space. Other works explore ways of more flexible and controllable text-driven image editing. Avrahami \etal~\cite{avrahami2022blended} design an algorithm in their Blended Diffusion framework to fuse noised versions of the input with the local text-guided diffusion latent, generating smooth transition. Similarly, GLIDE \cite{nichol2021glide} is also capable of handling text-guided inpainting after fine-tuning on this specific task.
Some methods give users more straightforward control on images. SDEdit \cite{meng2021sdedit} allows stroke painting in images, blending user input into the image by adding noise to the input and denoising through a stochastic differential equation. SDG \cite{liu2021more} injects Semantic Diffusion Guidance at each iteration of the generation and can provide multi-modal guidance.

While the versatility and generation quality of diffusion models have been repeatedly demonstrated, preserving object appearance remains a very challenging problem in image editing. Ruiz \etal~\cite{ruiz2022dreambooth} address this problem by optimizing their model using a reconstruction loss and a class-specific prior preservation loss. Though their approach can preserve the details of the object, multiple images of the same object are required to fine-tune their model on this target object. Kawar \etal~\cite{kawar2022imagic} handle this issue by interpolating the initial text embedding and the optimized embedding for reconstruction, but their method is mainly designed for applying edits on objects, not for generating the same object in different contexts.
In this paper, we propose the first generative object compositing framework based on diffusion models, which generates the harmonized and geometrically-corrected subject with a novel view and shadow. Furthermore, the characteristics of the original object is preserved in the synthesized composite image.

\section{Proposed Method}
\label{sec:method}

\begin{figure*}
  \centering
  \begin{subfigure}{\linewidth}
  \includegraphics[width=\textwidth]{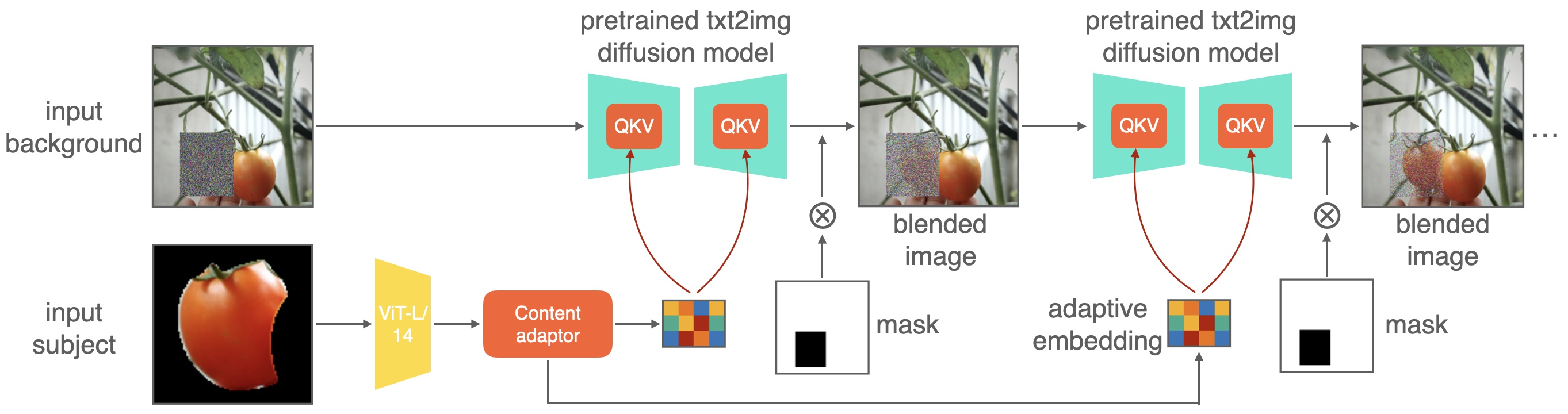}
  \end{subfigure}
  \vspace{-10pt}
  \caption{System pipeline. Our framework consists of a content adaptor and a generator (a pretrained text-to-image diffusion model). The input image $I_o$ is fed into a ViT and the adaptor which produces a descriptive embedding. At the same time the background image $I_{bg}$ is taken as input by the diffusion model. At each iteration during the denoising stage, we apply the mask $M$ on the generated image $I_{out}$, so that the generator only denoises the masked area $I_{out} \bigotimes M$.
  }
  \label{fig:main_pipeline}
\end{figure*}

We define the generative object compositing problem as: Given an input triplet $(I_o, I_{bg}, M)$ that consists of an object image $I_{o} \in \mathbb{R}^{H_s \times W_s \times 3}$, a background image $I_{bg} \in \mathbb{R}^{H_t \times W_t \times 3}$, and its associated binary mask $M \in \mathbb{R}^{H_t \times W_t \times 1}$ with the desired object location set to 0 and the rest to 1, the goal is to composite the input object into the masked area $I_{bg} \bigotimes M$. 
$M$ is considered as a soft constraint of the location and scale of the composited object.
The output generated image should look realistic, while the appearance of the object is preserved. 
Our problem setting is different from text-guided image generation and inpainting in that the condition input is a reference object image rather than a text prompt.
Inspired by the success of text-guided diffusion models \cite{rombach2022high, saharia2022photorealistic, ramesh2021zero},
which inject text conditioning into the diffusion architecture, we design our method for generative compositing to leverage such pretrained models.

An overview of our framework is shown in \cref{fig:main_pipeline}. It consists of an object image encoder extracting semantic features from the object and a conditional diffusion generator. 
To leverage the power of pretrained text-to-image diffusion models, we introduce a content adaptor that can bridge the gap between the object encoder and conditional generator by transforming a sequence of visual tokens to a sequence of text tokens to overcome the domain gap between image and text. This design further helps to preserve the object appearance.
We propose a two-stage training process:
in the first stage, the content adaptor is trained on large image/text pairs to maintain high-level semantics of the object; in the second stage, it is trained in the context of diffusion generator to encode key identity features of the object by encouraging the visual reconstruction of the object in the original image. Lastly, the generator module is fine-tuned on the embedding produced by the adaptor through cross attention blocks. All stages our trained in a self-supervised manner to avoid expensive annotation for obtaining object compositing training data.

\subsection{Generator}
\label{sec:generator}

As depicted in \cref{fig:main_pipeline}, we leverage a pretrained text-to-image diffusion model architecture and modify it for the compositing task by (i) introducing a mask in the input, and (ii) adjusting the U-Net input to contain the original background image outside the hole and noise inside the hole with mask blending.
In order to condition the model on the guidance embedding $E$, an attention mechanism is applied as:
\begin{equation}
  \text{Softmax}\left(\frac{(\boldsymbol{W}_{Q}E_{x}) (\boldsymbol{W}_{K}E)^{T}}{\sqrt{d}}\right)\boldsymbol{W}_{V}E = 
  \boldsymbol{A}\boldsymbol{V}
  \label{eq:cross_attn}
\end{equation}
where $E_{x}$ is an intermediate representation of the denoising autoencoder and $\boldsymbol{W}_{Q} \in \mathbb{R}^{d \times d_x}$, $\boldsymbol{W}_{K} \in \mathbb{R}^{d \times d_e}$ and $\boldsymbol{W}_{V} \in \mathbb{R}^{d \times d_e}$ are embedding matrices.
The background outside the mask area should be perfectly preserved for generative object compositing. Thus, we use the input mask $M$ for blending the background image $I_{bg}$ with the generated image $I_{out}$. As a result, the generator only denoises the masked area $I_{out} \bigotimes M $.
To use this model on our task, a straightforward way (a baseline) would be to apply image captioning on the object image and feeding the resulting caption directly to the diffusion model as the condition. However, text embedding cannot capture fine grained visual details. Therefore, in order to directly leverage the given object image, we introduce the content adaptor to transform the visual features from a pretrained visual encoder to text features (tokens) to use as conditioning for the generator.

\subsection{Content Adaptor}
\label{sec:translator}

\begin{figure}[t]
  \centering
   \includegraphics[width=\linewidth]{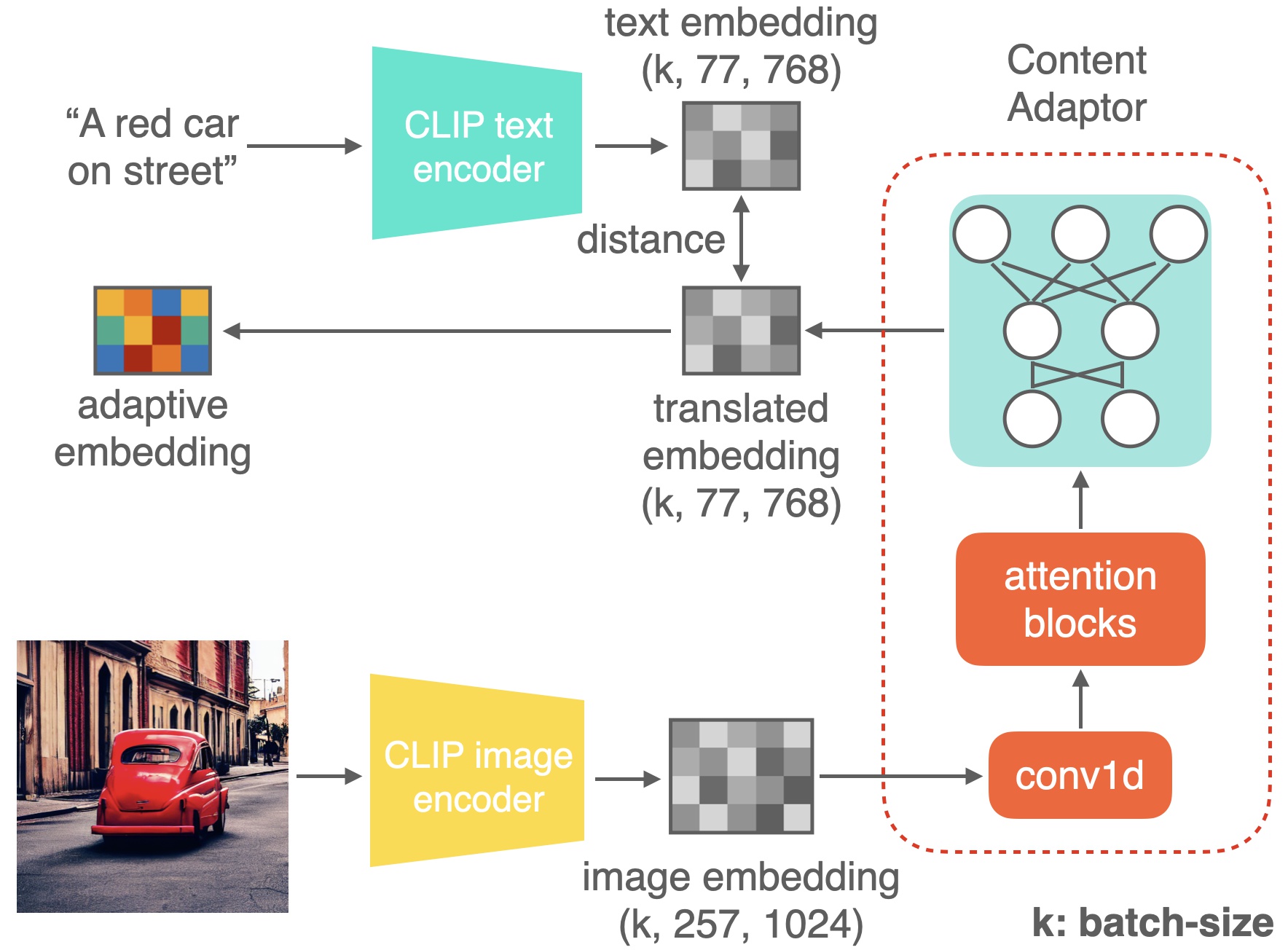}
   \vspace{-18pt}
   \caption{Structure of the Content Adaptor. In the first stage, it is trained on a large dataset of image-caption pairs to learn multi-modal sequential embeddings containing high-level semantics. In the second stage, it is fine-tuned under the diffusion framework to learn to encode identity features in \textit{adaptive embedding}. 
    }
   \label{fig:translator_pipeline}
\end{figure}

To prevent the loss of key identity information, we use an \textit{image encoder} instead of a text encoder to produce the embedding from the input object image. However, the image embedding cannot be effectively utilized by the diffusion model for two reasons:
\begin{itemize}[noitemsep, topsep=3pt]
    \item The image embedding $\widetilde{E}$ and the text embedding $E$ are from different domains. Since the diffusion model was trained on $E$, it cannot generate meaningful contents from image embedding sequence;
    \item A mismatch in the dimensions of $\widetilde{E} \in \mathbb{R}^{k \times 257 \times 1024}$ and $E \in \mathbb{R}^{k \times 77 \times 768}$, where $k$ is batch-size.
\end{itemize}

Therefore, based on the above observations, we develop a sequence-to-sequence translator architecture as shown in \cref{fig:translator_pipeline}. Given an image-caption pair as an input tuple $(I, t)$, we employ two pretrained ViT-L/14 encoders $C_{t}$, $ C_{i}$ from CLIP \cite{radford2021learning} to produce text embedding $E = C_{t}(t)$ and image embedding $\widetilde{E} = C_{i}(I)$, respectively. The adaptor $T$ consists of three components: a 1D convolutional layer, attention blocks \cite{vaswani2017attention} and an MLP, where the 1D convolution modifies the length of the embedding from 257 to 77, the MLP maps the embedding dimension from 1024 to 768, and the attention blocks bridge the gap between text domain and image domain. We design a two-stage optimization method to train this module, which is explained in \cref{sec:training}.

\subsection{Self-supervised Framework}
\label{sec:self_supervised}

There is no publicly available image compositing training dataset with annotations which is sufficient for training a diffusion model, and it is extremely challenging to manually annotate such data. Therefore, we propose a self-supervised training scheme and a synthetic data generation approach that simulate real-world scenarios. We also introduce a data augmentation method to enrich the training data as well as improve the robustness of our model.

\vspace{-5pt}
\subsubsection{Data Generation and Augmentation}

\label{sec:data}
\begin{figure}[t]
  \centering
   \includegraphics[width=\linewidth]{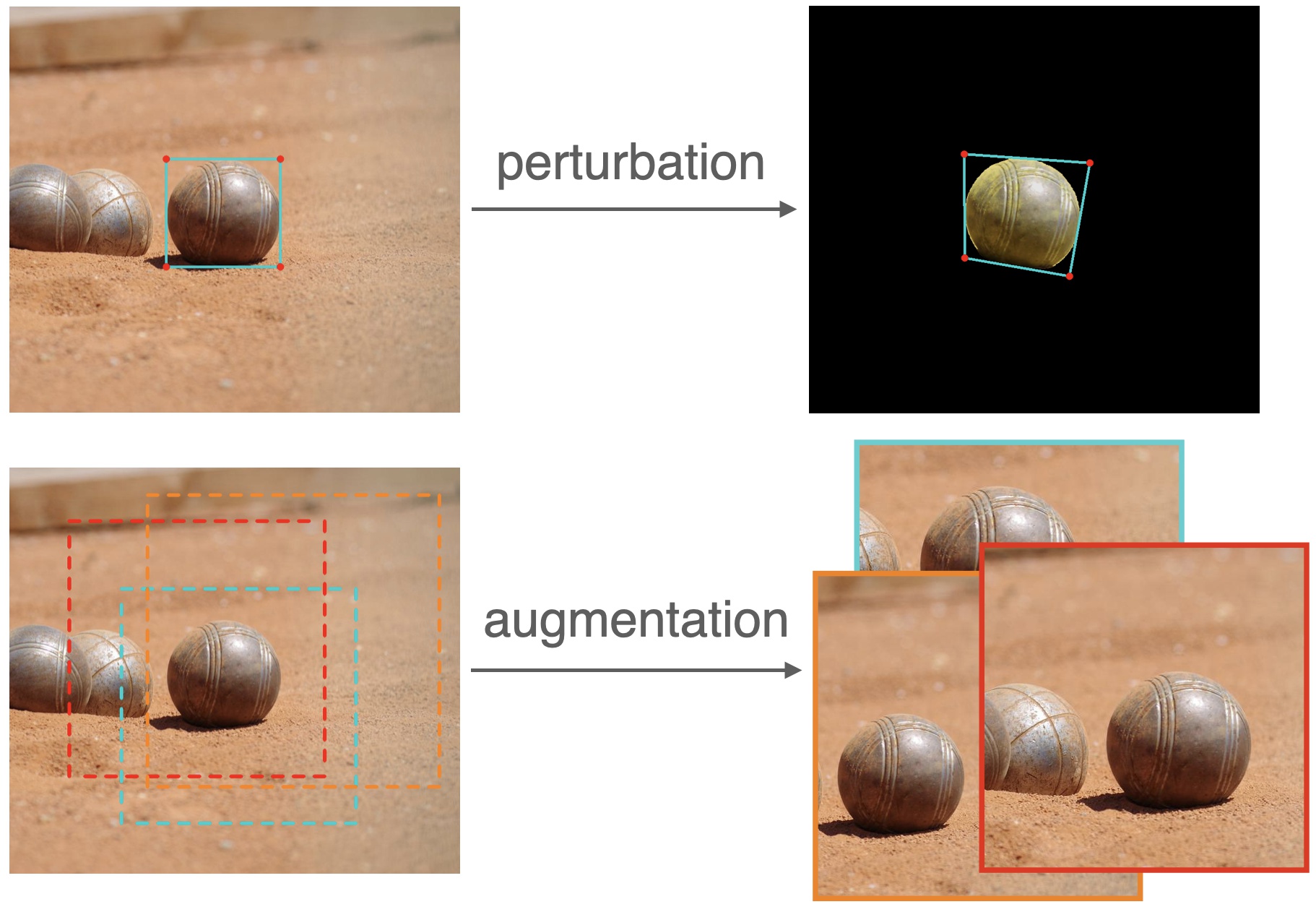}
   \vspace{-20pt}
   \caption{Illustration of our synthetic data generation and data augmentation scheme. The top row shows the data generation process including perspective warping, random rotation, and random color shifting. The original image is used as both the input background and ground truth, while the perturbed object is fed into the adaptor. The bottom row shows crop and shift augmentations, which helps to improve the generation quality and preserve object details.}
   \label{fig:data_gen}
\end{figure}

\textbf{Training data generation.} We collect our synthetic training data from Pixabay and use an object instance segmentation model \cite{lee2020centermask} to predict panoptic segmentation and classification labels. We first filter the dataset by removing objects with very small or large sizes. Then, we apply spatial and color perturbations to simulate many real use-case scenarios where the input image and background image have different scene geometry and lighting conditions.

The top row of \cref{fig:data_gen} illustrates this process. Inspired by \cite{detone2016deep}, we randomly perturb the four points of the object bounding box to apply projective transformation, followed by a random rotation within the range $[-\theta, \theta]$, where $\theta=20^{\circ}$, and color perturbation. The segmentation mask (perturbed in the same way as the image) is used to extract the object.

This data synthesis pipeline is fully controllable and can also be employed as data augmentation during training. Another key advantage is that it is free of manual labeling, since the original image is used as the ground truth.
We use the bounding box as the mask as it not only fully covers the object, but also extends to its neighboring area (providing room for shadow generation). We find that it is flexible enough for the model to apply spatial transformations, synthesize novel views and generate shadows and reflection.

\vspace{3pt}
\noindent \textbf{Real-world evaluation data.}
Given that there is no existing dataset specifically for our task, we manually collect a new dataset, closely simulating real-world use cases, as an evaluation benchmark for object compositing.
The dataset consists of 503 pairs of common objects (including both rigid and non-rigid objects such as vehicles and wildlife) and diverse background images (covering both indoor and outdoor scenes). It also contains challenging cases where there is a large mismatch of lighting conditions or viewpoints between foreground and background.
The dataset images are collected from Pixabay.
The labeling procedure closely simulate the real scenarios where the input object is placed at a target location in the background image and then scaled at the user's will. 
The compositing region is determined as a loose bounding box around the object. 
More details of the labeling tool are provided in the supplementary material.

\vspace{3pt}
\noindent \textbf{Data augmentation.}
Inspired by \cite{lee2021vision}, we introduce random shift and crop augmentations during training, while ensuring that the foreground object is always contained in the crop window. This process is illustrated in the bottom row of \cref{fig:data_gen}.
Applying this augmentation method for both training and inference results in a notable improvement in the realism of the generated results. Quantitative results are provided in \cref{sec:as}. 

\vspace{-5pt}
\subsubsection{Training}
\label{sec:training}

\textbf{Content adaptor pretraining.}
We first pretrain the content adaptor to keep the semantics of the object by mapping the image embedding to text embedding. At this first stage, we optimize the content adaptor on a sequence-to-sequence translation task, which learns to project the image embedding into a multi-modal space. During training, $C_t$, $C_i$ are frozen, and the translator is trained on 3,203,338 image-caption pairs from a filtered LAION dataset \cite{schuhmann2021laion}.

Given the input image embedding $\widetilde{E}$, we use the text embedding $E$ as target, the objective function of this translation task is defined as:
\begin{equation}
  \mathcal{L}_{dist} = \lVert T(\tilde{E}) - E \rVert_{1},
  \label{eq:translate_loss}
\end{equation}
where $T(\cdot)$ is the content adaptor.

However, the multi-modal embedding obtained solely from this first stage optimization mostly carry high-level semantics of the input image without much of texture details. Hence, we further refine $\widetilde{E}$ to obtain $\hat{E}$, the adaptive embedding, for better appearance preservation.


\vspace{3pt}
\noindent \textbf{Content adaptor fine-tuning.}
We further optimize the content adaptor to produce an adaptive embedding which maintains instance-level properties of the object.
After pretraining, we insert the content adaptor to the diffusion framework by feeding the adaptive embedding as context to the attention blocks. Then, the diffusion model is frozen and the adaptor is trained using:
\begin{equation}
  \mathcal{L}_{adapt} = \mathbb{E}_{T, \epsilon \sim \mathcal{N}(0, 1)} \lVert \epsilon - \epsilon_{\theta}(I_t \circ M, t, T(\widetilde{E})) \rVert_{2}^{2},
  \label{eq:opt_loss}
\end{equation}
where the content adaptor $T(\cdot)$ is optimized. It is trained on our synthetic dataset filtered from Pixabay, containing 467,280 foreground and background image pairs for training and 25,960 pairs for validation.

\vspace{3pt}
\noindent \textbf{Generator fine-tuning.}
After the aforementioned two-stage training of the content adaptor is completed, we freeze the content adaptor and train the generator module after initializing the text-to-image diffusion model with pretrained weights. Crop and shift augmentations are applied in this process. Based on the Latent Diffusion model \cite{rombach2022high}, the objective loss in our generator module is defined by:
\begin{equation}
  \mathcal{L}_{gen} = \mathbb{E}_{\hat{E}, \epsilon \sim \mathcal{N}(0, 1)} \lVert \epsilon - \epsilon_{\theta}(I_t \circ M, t, \hat{E}) \rVert_{2}^{2},
  \label{eq:gen_loss}
\end{equation}
where $I$ is the input image, $I_t$ is a noisy version of $I$ at timestep $t$, and $\epsilon_{\theta}$ denotes the denoising model which is optimized. In order to adapt the text-guided generation task to the setting of image compositing, we apply the input mask on the image at every time-step.

\section{Experiments}
\label{sec:exp}

\subsection{Training Details}
\label{sec:imp}

The first training stage of the content adaptor takes 15 epochs with a learning rate of $10^{-4}$ and batch size of 2048. The image size processed by ViT is 224 $\times$ 224. To set up the pipeline, we employed the pretrained image encoder and text encoder from CLIP~\cite{radford2021learning}. 
The second training stage of the content adaptor takes 13 epochs with the learning rate of $2\times10^{-5}$ and batch size of 512. We resize the images to 512 $\times$ 512 and adapt the pretrained Stable Diffusion model~\cite{rombach2022high} to our task.
After the above process, the generator is trained for 20 epochs with the learning rate of $4\times10^{-5}$ and batch size of 576. The input image size is the same as that in the second training stage.
All training stages are conducted on 8 A100 GPUs with Adam optimizer.

\subsection{Quantitative Evaluation}

\begin{figure}[t]
  \centering
   \includegraphics[width=\linewidth]{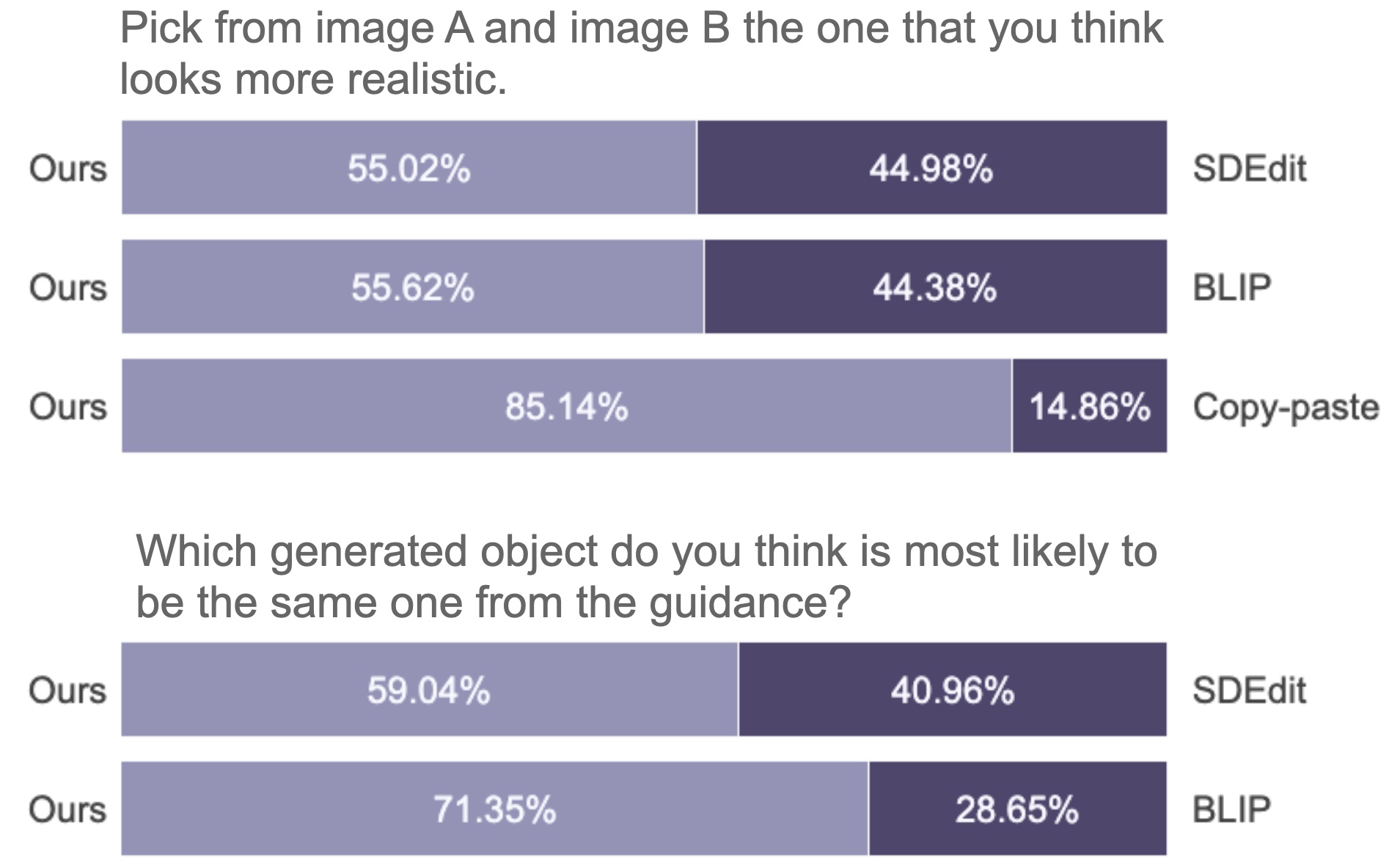}
   \vspace{-18pt}
   \caption{User study results. We conduct side-by-side comparisons between our method and one of baseline methods to quantify the generation quality in terms of realism and appearance preservation. The results show that our method outperforms baselines.}
   \vspace{-5pt}
   \label{fig:user_study}
\end{figure}

\begin{figure*}
  \centering
  \begin{subfigure}{\linewidth}
  \includegraphics[width=\textwidth]{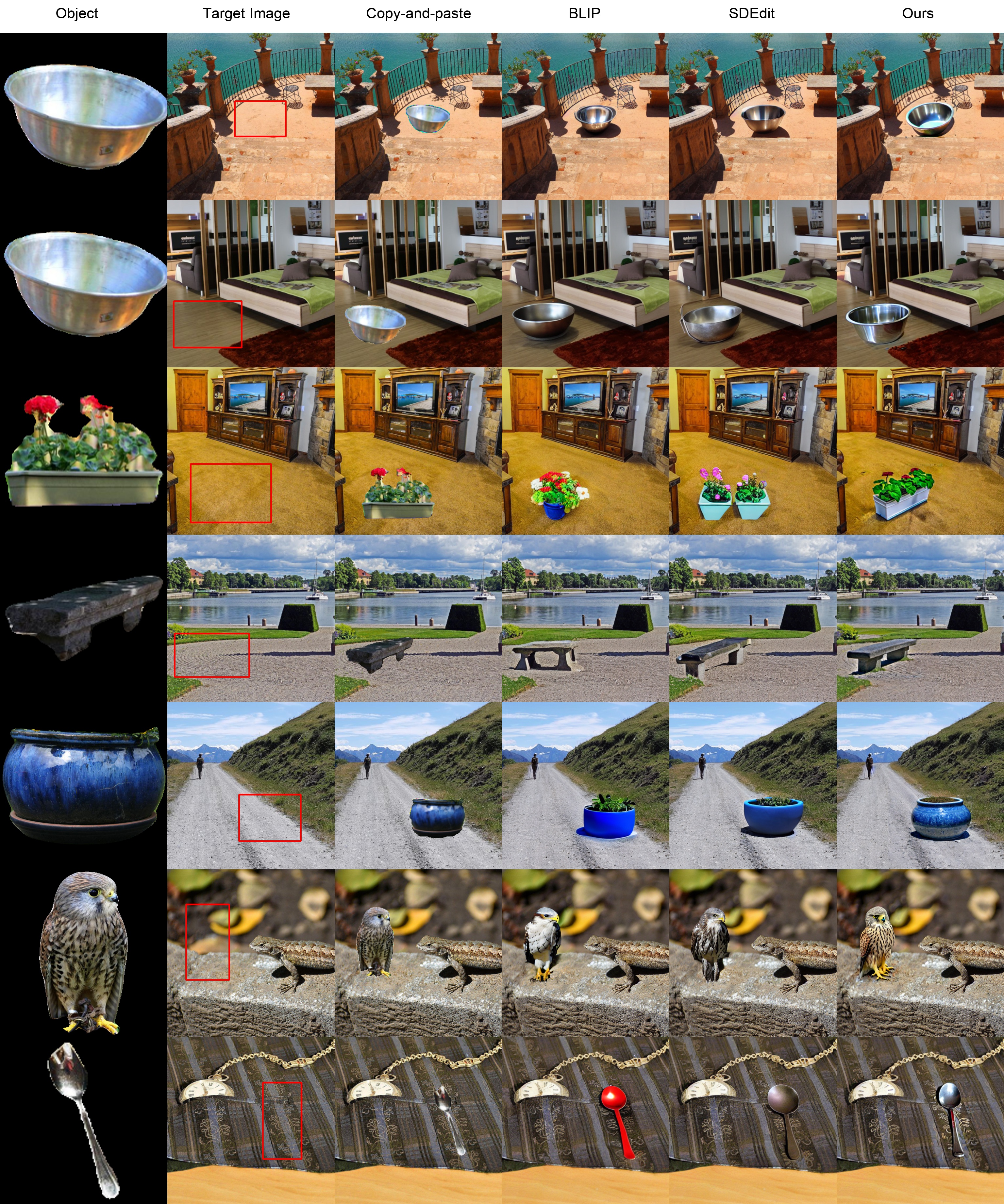}
  \end{subfigure}
  \caption{Qualitative comparison on the real-world evaluation dataset. Our method can appropriately correct the viewpoint and geometry while harmonizing the object with the new background, generating realistic composite results. Moreover, lighting and shadows are naturally adjusted in the composite image. Compared to baselines using text guidance, our method can better preserve the object appearance.}
  \label{fig:qual}
\end{figure*}

\begin{table*}[t]
  \centering
  \small
  \begin{tabular}{lllllll}
    \toprule
    Adaptor & Optimization & Augmentation & BLIP \cite{li2022blip} & FID $\downarrow$ & CLIP text score $\uparrow$ & CLIP image score $\uparrow$ \\
    \midrule
    \cmark & \cmark & \cmark & \xmark & \textbf{15.4331} & \textbf{29.8594} & \textbf{97.0625} \\
    \xmark & \cmark & \cmark & \xmark & 18.8254 & 29.8281 & 96.0000 \\
    \cmark & \xmark & \cmark & \xmark & 16.1381 & 29.7969 & 96.8125 \\
    \cmark & \cmark & \xmark & \xmark & 18.3698 & 29.7813 & 96.1875 \\
    \xmark & \xmark & \cmark & \cmark & 17.9000 & 29.7188 & 95.8125 \\
    \bottomrule
  \end{tabular}
  \vspace{-5pt}
  \caption{Ablation study.
  We evaluate the effectiveness of the components listed in the first four columns: 1) whether the content adaptor is used; 2) whether the adaptor is optimized for appearance preservation; 3) whether crop augmentation is utilized in training; 4) if the adaptor module is replaced by BLIP to predict text embedding from an image. We use FID to assess the generation fidelity and modified CLIP score (explained in \cref{sec:as}) to measure similarity of the guidance and the predicted image.
  }
  \label{tab:as_table}
\end{table*}

To obtain quantitative results on a real dataset that simulates real-world use cases, where users drag and drop an object to a background image, we conduct a user study to compare our method to three baselines, \ie, copy-and-paste, pretrained stable diffusion model fine-tuned on BLIP~\cite{li2022blip}, and a joint stable diffusion model integrating both BLIP and SDEdit~\cite{meng2021sdedit}. Given the fact that there is no existing diffusion-based method addressing the same problem, we train the baselines on the same pretrained diffusion model using the same self-supervised training scheme and dataset as ours.
To the best of our knowledge, after adapting to our setting, they are the closest baselines to our task. 
In the BLIP baseline, the content adaptor module is replaced by BLIP, a state-of-the-art captioning model. The text embedding obtained from the BLIP caption is fed into the generator module. For the SDEdit baseline, we keep the BLIP model which will significantly improve the generation fidelity of a pretrained diffusion model.
We set the noise strength to 0.65 in SDEdit, which strikes the balance between realism and faithfulness.

We perform a user study using our real testing dataset, which consists of 503 object-background pairs. 
For each example, we set up a side-by-side comparison between the results from one of the baselines and our method, respectively. The subject will be asked two questions: 1) which is more realistic and 2) which generated object is more close to the given object? 
We collected 1,494 votes from more than 170 users. 
As shown in \cref{fig:user_study}, our method outperforms the other baselines for both questions.
The higher preference rates demonstrate the effectiveness of the content adaptor and data augmentation method in improving generation fidelity. It also shows that our content adaptor module is capable of maintaining details and attributes of the input objects, while not being constrained by the input when harmonizing objects to the background. In contrast, although SDEdit is better at preserving appearance than BLIP, it is constrained by the original pose, shape, and other attributes of the object, making the generated object less coherent with the background.

\subsection{Qualitative Evaluation}
In \cref{fig:qual}, we compare our method to various baselines.
In the first two rows of \cref{fig:qual}, we place the same object in two scenes with notable differences in the geometry. Results show that our model is capable of correcting geometric inconsistencies while preserving the characteristics of the object. The limitations of the baselines are also shown in \cref{fig:qual}: BLIP often fails to preserve the appearance; SDEdit preserves texture and pose better than the first baseline, but often cannot make appropriate geometry transformations.

\subsection{Ablation Study}
\label{sec:as}
In \cref{tab:as_table}, we demonstrate the effectiveness of each key component in our ObjectStitch framework by ablating each of them.
We use FID\cite{heusel2017gans} and a modified CLIP~\cite{hessel2021clipscore, radford2021learning} score, which measure the fidelity and semantic matching performance between given and generated objects. Instead of directly adopting the CLIP score, which semantically matches the prompt and the image, we modify it to \textit{CLIP text score} and \textit{CLIP image score} that are defined as follows:
\begin{flalign}
    \mathcal{C}_{txt} & = E\left[s \cdot f(I_{pred}) \cdot g\left(B(I_{gt})\right)\right], \\
    \mathcal{C}_{img} & = E\left[s \cdot f(I_{pred}) \cdot f(I_{gt})\right],
    \label{eq:clip_score}
\end{flalign}
where $B(\cdot)$ is pretrained BLIP~\cite{li2022blip}, and $s$ is a logit scale.

For the ablation models, we first remove the content adaptor to evaluate its effect. In order to keep the embedding dimension fed to the generator, we removed the attention blocks from the content adaptor. We also drop the bias and activation function of the last linear layer, such that in the adaptation process, the model will only learn a non-adaptive embedding dictionary. Without the attention layers, the model has difficulty bridging the domain gap between text and image, resulting in the degradation of generation fidelity as shown in the second row of \cref{tab:as_table}.

The third row in \cref{tab:as_table} ablates the second training stage of the content adaptor by simply skipping it. Without this process, the model fails to learn descriptive embedding, which encodes instance-level features of the guidance object, leading to a drop in appearance preservation.

Data augmentation is another key element for better performance, \ie, cropping and shift augmentation
increases the mask area, thus improving the generation quality in detail. The fourth row in \cref{tab:as_table} illustrates that with the absence of data augmentation, there is a notable drop in realism, and the generation carries less accurate details.

To further assess the importance of the content adaptor module, we replace it with a pretrained BLIP model, which transfers the reference image to a text embedding. As shown in the last row in \cref{tab:as_table}, this results in the lowest CLIP image score, indicating that high-level semantics represented by text are not sufficient for maintaining object identity, highlighting the importance of our content adaptor.

\section{Conclusion and Future Work}
\label{sec:con}
We propose the first diffusion-based approach to tackle generative object compositing. In contrast to the traditional compositing pipeline which requires many steps including accurate segmentation, harmonization, geometry correction, view synthesis and shadow generation, our method can directly achieve realistic composite results. We construct our model by modifying foundational text-guided image generation framework and introducing a novel content adaptor module. Furthermore, we introduce a fully self-supervised framework to train our model with synthetic data generation and data augmentation. We show that our method outperforms the baseline methods through user study results on real-world examples.

Since this is the first attempt toward a challenging task, the proposed method has several limitations. It lacks control in appearance preservation of the synthesized object with respect to the the provided reference object image. A possible idea to improve this aspect is to train the visual encoder on pairs of images of the same object or train it jointly with the rest of the framework. Another key limitation is caused by masking the output image, which prohibits our model from generating global effects, \ie shadow can only be synthesized within the mask. To address this issue, we may need to synthesize complete background images and pair them with objects, and train with a conditional generation framework.


{\small
\bibliographystyle{ieee_fullname}
\bibliography{PaperForreview}

\begin{thebibliography}{10}\itemsep=-1pt

\bibitem{avrahami2022blended}
Omri Avrahami, Dani Lischinski, and Ohad Fried.
\newblock Blended diffusion for text-driven editing of natural images.
\newblock In {\em Proceedings of the IEEE/CVF Conference on Computer Vision and
  Pattern Recognition}, pages 18208--18218, 2022.

\bibitem{azadi2020compositional}
Samaneh Azadi, Deepak Pathak, Sayna Ebrahimi, and Trevor Darrell.
\newblock Compositional gan: Learning image-conditional binary composition.
\newblock {\em International Journal of Computer Vision}, 128(10):2570--2585,
  2020.

\bibitem{balaji2022ediffi}
Yogesh Balaji, Seungjun Nah, Xun Huang, Arash Vahdat, Jiaming Song, Karsten
  Kreis, Miika Aittala, Timo Aila, Samuli Laine, Bryan Catanzaro, et~al.
\newblock ediffi: Text-to-image diffusion models with an ensemble of expert
  denoisers.
\newblock {\em arXiv preprint arXiv:2211.01324}, 2022.

\bibitem{brock2018large}
Andrew Brock, Jeff Donahue, and Karen Simonyan.
\newblock Large scale gan training for high fidelity natural image synthesis.
\newblock {\em arXiv preprint arXiv:1809.11096}, 2018.

\bibitem{chen2019toward}
Bor-Chun Chen and Andrew Kae.
\newblock Toward realistic image compositing with adversarial learning.
\newblock In {\em Proceedings of the IEEE/CVF Conference on Computer Vision and
  Pattern Recognition}, pages 8415--8424, 2019.

\bibitem{cong2021bargainnet}
Wenyan Cong, Li Niu, Jianfu Zhang, Jing Liang, and Liqing Zhang.
\newblock Bargainnet: Background-guided domain translation for image
  harmonization.
\newblock In {\em 2021 IEEE International Conference on Multimedia and Expo
  (ICME)}, pages 1--6. IEEE, 2021.

\bibitem{cong2020dovenet}
Wenyan Cong, Jianfu Zhang, Li Niu, Liu Liu, Zhixin Ling, Weiyuan Li, and Liqing
  Zhang.
\newblock Dovenet: Deep image harmonization via domain verification.
\newblock In {\em Proceedings of the IEEE/CVF Conference on Computer Vision and
  Pattern Recognition}, pages 8394--8403, 2020.

\bibitem{detone2016deep}
Daniel DeTone, Tomasz Malisiewicz, and Andrew Rabinovich.
\newblock Deep image homography estimation.
\newblock {\em arXiv preprint arXiv:1606.03798}, 2016.

\bibitem{feng2022ernie}
Zhida Feng, Zhenyu Zhang, Xintong Yu, Yewei Fang, Lanxin Li, Xuyi Chen, Yuxiang
  Lu, Jiaxiang Liu, Weichong Yin, Shikun Feng, et~al.
\newblock Ernie-vilg 2.0: Improving text-to-image diffusion model with
  knowledge-enhanced mixture-of-denoising-experts.
\newblock {\em arXiv preprint arXiv:2210.15257}, 2022.

\bibitem{goodfellow2020generative}
Ian Goodfellow, Jean Pouget-Abadie, Mehdi Mirza, Bing Xu, David Warde-Farley,
  Sherjil Ozair, Aaron Courville, and Yoshua Bengio.
\newblock Generative adversarial networks.
\newblock {\em Communications of the ACM}, 63(11):139--144, 2020.

\bibitem{hessel2021clipscore}
Jack Hessel, Ari Holtzman, Maxwell Forbes, Ronan~Le Bras, and Yejin Choi.
\newblock Clipscore: A reference-free evaluation metric for image captioning.
\newblock {\em arXiv preprint arXiv:2104.08718}, 2021.

\bibitem{heusel2017gans}
Martin Heusel, Hubert Ramsauer, Thomas Unterthiner, Bernhard Nessler, and Sepp
  Hochreiter.
\newblock Gans trained by a two time-scale update rule converge to a local nash
  equilibrium.
\newblock {\em Advances in neural information processing systems}, 30, 2017.

\bibitem{ho2020denoising}
Jonathan Ho, Ajay Jain, and Pieter Abbeel.
\newblock Denoising diffusion probabilistic models.
\newblock {\em Advances in Neural Information Processing Systems},
  33:6840--6851, 2020.

\bibitem{hold2017deep}
Yannick Hold-Geoffroy, Kalyan Sunkavalli, Sunil Hadap, Emiliano Gambaretto, and
  Jean-Fran{\c{c}}ois Lalonde.
\newblock Deep outdoor illumination estimation.
\newblock In {\em Proceedings of the IEEE conference on computer vision and
  pattern recognition}, pages 7312--7321, 2017.

\bibitem{hong2022shadow}
Yan Hong, Li Niu, and Jianfu Zhang.
\newblock Shadow generation for composite image in real-world scenes.
\newblock In {\em Proceedings of the AAAI Conference on Artificial
  Intelligence}, volume~36, pages 914--922, 2022.

\bibitem{isola2017image}
Phillip Isola, Jun-Yan Zhu, Tinghui Zhou, and Alexei~A Efros.
\newblock Image-to-image translation with conditional adversarial networks.
\newblock In {\em Proceedings of the IEEE conference on computer vision and
  pattern recognition}, pages 1125--1134, 2017.

\bibitem{jaderberg2015spatial}
Max Jaderberg, Karen Simonyan, Andrew Zisserman, et~al.
\newblock Spatial transformer networks.
\newblock {\em Advances in neural information processing systems}, 28, 2015.

\bibitem{jiang2021ssh}
Yifan Jiang, He Zhang, Jianming Zhang, Yilin Wang, Zhe Lin, Kalyan Sunkavalli,
  Simon Chen, Sohrab Amirghodsi, Sarah Kong, and Zhangyang Wang.
\newblock Ssh: A self-supervised framework for image harmonization.
\newblock In {\em Proceedings of the IEEE/CVF International Conference on
  Computer Vision}, pages 4832--4841, 2021.

\bibitem{karras2019style}
Tero Karras, Samuli Laine, and Timo Aila.
\newblock A style-based generator architecture for generative adversarial
  networks.
\newblock In {\em Proceedings of the IEEE/CVF conference on computer vision and
  pattern recognition}, pages 4401--4410, 2019.

\bibitem{karsch2011rendering}
Kevin Karsch, Varsha Hedau, David Forsyth, and Derek Hoiem.
\newblock Rendering synthetic objects into legacy photographs.
\newblock {\em ACM Transactions on Graphics (TOG)}, 30(6):1--12, 2011.

\bibitem{karsch2014automatic}
Kevin Karsch, Kalyan Sunkavalli, Sunil Hadap, Nathan Carr, Hailin Jin, Rafael
  Fonte, Michael Sittig, and David Forsyth.
\newblock Automatic scene inference for 3d object compositing.
\newblock {\em ACM Transactions on Graphics (TOG)}, 33(3):1--15, 2014.

\bibitem{kawar2022imagic}
Bahjat Kawar, Shiran Zada, Oran Lang, Omer Tov, Huiwen Chang, Tali Dekel, Inbar
  Mosseri, and Michal Irani.
\newblock Imagic: Text-based real image editing with diffusion models.
\newblock {\em arXiv preprint arXiv:2210.09276}, 2022.

\bibitem{lalonde2007using}
Jean-Francois Lalonde and Alexei~A Efros.
\newblock Using color compatibility for assessing image realism.
\newblock In {\em 2007 IEEE 11th International Conference on Computer Vision},
  pages 1--8. IEEE, 2007.

\bibitem{lee2021vision}
Seung~Hoon Lee, Seunghyun Lee, and Byung~Cheol Song.
\newblock Vision transformer for small-size datasets.
\newblock {\em arXiv preprint arXiv:2112.13492}, 2021.

\bibitem{lee2020centermask}
Youngwan Lee and Jongyoul Park.
\newblock Centermask: Real-time anchor-free instance segmentation.
\newblock In {\em Proceedings of the IEEE/CVF conference on computer vision and
  pattern recognition}, pages 13906--13915, 2020.

\bibitem{li2022blip}
Junnan Li, Dongxu Li, Caiming Xiong, and Steven Hoi.
\newblock Blip: Bootstrapping language-image pre-training for unified
  vision-language understanding and generation.
\newblock {\em arXiv preprint arXiv:2201.12086}, 2022.

\bibitem{lin2018st}
Chen-Hsuan Lin, Ersin Yumer, Oliver Wang, Eli Shechtman, and Simon Lucey.
\newblock St-gan: Spatial transformer generative adversarial networks for image
  compositing.
\newblock In {\em Proceedings of the IEEE Conference on Computer Vision and
  Pattern Recognition}, pages 9455--9464, 2018.

\bibitem{liu2020arshadowgan}
Daquan Liu, Chengjiang Long, Hongpan Zhang, Hanning Yu, Xinzhi Dong, and
  Chunxia Xiao.
\newblock Arshadowgan: Shadow generative adversarial network for augmented
  reality in single light scenes.
\newblock In {\em Proceedings of the IEEE/CVF Conference on Computer Vision and
  Pattern Recognition}, pages 8139--8148, 2020.

\bibitem{liu2021more}
Xihui Liu, Dong~Huk Park, Samaneh Azadi, Gong Zhang, Arman Chopikyan, Yuxiao
  Hu, Humphrey Shi, Anna Rohrbach, and Trevor Darrell.
\newblock More control for free! image synthesis with semantic diffusion
  guidance.
\newblock {\em arXiv preprint arXiv:2112.05744}, 2021.

\bibitem{luo2021diffusion}
Shitong Luo and Wei Hu.
\newblock Diffusion probabilistic models for 3d point cloud generation.
\newblock In {\em Proceedings of the IEEE/CVF Conference on Computer Vision and
  Pattern Recognition}, pages 2837--2845, 2021.

\bibitem{meng2021sdedit}
Chenlin Meng, Yang Song, Jiaming Song, Jiajun Wu, Jun-Yan Zhu, and Stefano
  Ermon.
\newblock Sdedit: Image synthesis and editing with stochastic differential
  equations.
\newblock {\em arXiv preprint arXiv:2108.01073}, 2021.

\bibitem{nichol2021glide}
Alex Nichol, Prafulla Dhariwal, Aditya Ramesh, Pranav Shyam, Pamela Mishkin,
  Bob McGrew, Ilya Sutskever, and Mark Chen.
\newblock Glide: Towards photorealistic image generation and editing with
  text-guided diffusion models.
\newblock {\em arXiv preprint arXiv:2112.10741}, 2021.

\bibitem{perez2003poisson}
Patrick P{\'e}rez, Michel Gangnet, and Andrew Blake.
\newblock Poisson image editing.
\newblock In {\em ACM SIGGRAPH 2003 Papers}, pages 313--318. 2003.

\bibitem{poole2022dreamfusion}
Ben Poole, Ajay Jain, Jonathan~T Barron, and Ben Mildenhall.
\newblock Dreamfusion: Text-to-3d using 2d diffusion.
\newblock {\em arXiv preprint arXiv:2209.14988}, 2022.

\bibitem{radford2021learning}
Alec Radford, Jong~Wook Kim, Chris Hallacy, Aditya Ramesh, Gabriel Goh,
  Sandhini Agarwal, Girish Sastry, Amanda Askell, Pamela Mishkin, Jack Clark,
  et~al.
\newblock Learning transferable visual models from natural language
  supervision.
\newblock In {\em International Conference on Machine Learning}, pages
  8748--8763. PMLR, 2021.

\bibitem{ramesh2021zero}
Aditya Ramesh, Mikhail Pavlov, Gabriel Goh, Scott Gray, Chelsea Voss, Alec
  Radford, Mark Chen, and Ilya Sutskever.
\newblock Zero-shot text-to-image generation.
\newblock In {\em International Conference on Machine Learning}, pages
  8821--8831. PMLR, 2021.

\bibitem{rombach2022high}
Robin Rombach, Andreas Blattmann, Dominik Lorenz, Patrick Esser, and Bj{\"o}rn
  Ommer.
\newblock High-resolution image synthesis with latent diffusion models.
\newblock In {\em Proceedings of the IEEE/CVF Conference on Computer Vision and
  Pattern Recognition}, pages 10684--10695, 2022.

\bibitem{ruiz2022dreambooth}
Nataniel Ruiz, Yuanzhen Li, Varun Jampani, Yael Pritch, Michael Rubinstein, and
  Kfir Aberman.
\newblock Dreambooth: Fine tuning text-to-image diffusion models for
  subject-driven generation.
\newblock {\em arXiv preprint arXiv:2208.12242}, 2022.

\bibitem{saharia2022photorealistic}
Chitwan Saharia, William Chan, Saurabh Saxena, Lala Li, Jay Whang, Emily
  Denton, Seyed Kamyar~Seyed Ghasemipour, Burcu~Karagol Ayan, S~Sara Mahdavi,
  Rapha~Gontijo Lopes, et~al.
\newblock Photorealistic text-to-image diffusion models with deep language
  understanding.
\newblock {\em arXiv preprint arXiv:2205.11487}, 2022.

\bibitem{schuhmann2021laion}
Christoph Schuhmann, Richard Vencu, Romain Beaumont, Robert Kaczmarczyk,
  Clayton Mullis, Aarush Katta, Theo Coombes, Jenia Jitsev, and Aran
  Komatsuzaki.
\newblock Laion-400m: Open dataset of clip-filtered 400 million image-text
  pairs.
\newblock {\em arXiv preprint arXiv:2111.02114}, 2021.

\bibitem{sheng2022controllable}
Yichen Sheng, Yifan Liu, Jianming Zhang, Wei Yin, A~Cengiz Oztireli, He Zhang,
  Zhe Lin, Eli Shechtman, and Bedrich Benes.
\newblock Controllable shadow generation using pixel height maps.
\newblock In {\em European Conference on Computer Vision}, pages 240--256.
  Springer, 2022.

\bibitem{sheng2021ssn}
Yichen Sheng, Jianming Zhang, and Bedrich Benes.
\newblock Ssn: Soft shadow network for image compositing.
\newblock In {\em Proceedings of the IEEE/CVF Conference on Computer Vision and
  Pattern Recognition}, pages 4380--4390, 2021.

\bibitem{sohl2015deep}
Jascha Sohl-Dickstein, Eric Weiss, Niru Maheswaranathan, and Surya Ganguli.
\newblock Deep unsupervised learning using nonequilibrium thermodynamics.
\newblock In {\em International Conference on Machine Learning}, pages
  2256--2265. PMLR, 2015.

\bibitem{song2019generative}
Yang Song and Stefano Ermon.
\newblock Generative modeling by estimating gradients of the data distribution.
\newblock {\em Advances in Neural Information Processing Systems}, 32, 2019.

\bibitem{sunkavalli2010multi}
Kalyan Sunkavalli, Micah~K Johnson, Wojciech Matusik, and Hanspeter Pfister.
\newblock Multi-scale image harmonization.
\newblock {\em ACM Transactions on Graphics (TOG)}, 29(4):1--10, 2010.

\bibitem{tsai2017deep}
Yi-Hsuan Tsai, Xiaohui Shen, Zhe Lin, Kalyan Sunkavalli, Xin Lu, and Ming-Hsuan
  Yang.
\newblock Deep image harmonization.
\newblock In {\em Proceedings of the IEEE Conference on Computer Vision and
  Pattern Recognition}, pages 3789--3797, 2017.

\bibitem{vaswani2017attention}
Ashish Vaswani, Noam Shazeer, Niki Parmar, Jakob Uszkoreit, Llion Jones,
  Aidan~N Gomez, {\L}ukasz Kaiser, and Illia Polosukhin.
\newblock Attention is all you need.
\newblock {\em Advances in neural information processing systems}, 30, 2017.

\bibitem{xu2017deep}
Ning Xu, Brian Price, Scott Cohen, and Thomas Huang.
\newblock Deep image matting.
\newblock In {\em Proceedings of the IEEE conference on computer vision and
  pattern recognition}, pages 2970--2979, 2017.

\bibitem{xue2022dccf}
Ben Xue, Shenghui Ran, Quan Chen, Rongfei Jia, Binqiang Zhao, and Xing Tang.
\newblock Dccf: Deep comprehensible color filter learning framework for
  high-resolution image harmonization.
\newblock {\em arXiv preprint arXiv:2207.04788}, 2022.

\bibitem{zhang2018unreasonable}
Richard Zhang, Phillip Isola, Alexei~A Efros, Eli Shechtman, and Oliver Wang.
\newblock The unreasonable effectiveness of deep features as a perceptual
  metric.
\newblock In {\em Proceedings of the IEEE conference on computer vision and
  pattern recognition}, pages 586--595, 2018.

\bibitem{zhou2019deep}
Hao Zhou, Sunil Hadap, Kalyan Sunkavalli, and David~W Jacobs.
\newblock Deep single-image portrait relighting.
\newblock In {\em Proceedings of the IEEE/CVF International Conference on
  Computer Vision}, pages 7194--7202, 2019.

\end{thebibliography}
}

\vfil \break \clearpage

\begin{appendices}


\section{Overview}
The following sections are covered in this supplementary material to support our main paper:

\begin{itemize}
    \item Quantitative results of our method and the baselines;
    \item Real-world dataset collection;
    \item Our model's robustness against low-quality images;
    \item Additional qualitative examples;
    \item Architecture details and hyperparameters.
\end{itemize}


\section{Quantitative Results}
\label{sec:quan}

As mentioned in the paper, to demonstrate the realism and faithfulness of our model based on human perception, we conduct a user study with a real-world dataset and collect quantitative results.

To further support our conclusion, we prepared a synthetic test dataset based on Pixabay Dataset which is generated in a similar way to how we prepared the training dataset for our framework. The main difference is that we apply larger spatial perturbations on the input object. More specifically, the object is randomly rotated within the range $ [-\theta, \theta] $ where $\theta=40^{\circ}$. We apply this change to better evaluate the model's ability to correct large geometric inconsistencies.

We test our model and two baseline methods (BLIP \cite{li2022blip} and SDEdit \cite{meng2021sdedit}) on 1500 images randomly chosen from the synthetic test dataset. The baselines are trained on the same pretrained diffusion model, synthetic dataset, and using the same data augmentation method as ours. For SDEdit we use a noise strength of 0.85, enabling it to apply larger spatial transformations, which better adapts to the synthetic test dataset. In \cref{tab:quan}, we use FID \cite{heusel2017gans} as a measurement of fidelity, and LPIPS \cite{zhang2018unreasonable} to measure the feature distance between the generation and ground truth. We also employ a modified CLIP score \cite{hessel2021clipscore, radford2021learning} to measure semantic similarity between the given and generated object. The column \textit{Crop} indicates whether we compare the performance with a cropped square patch that covers the generated area. Focusing on the cropped region, we can better evaluate the generation quality; using full image, we can assess the matching performance between the generated area and the background. As shown in this table, our model achieves in all cases the best performance in fidelity and preservation.
Despite using a combination of metrics well suited to our task, we observe there are still limitations in these evaluation methods. For example, they cannot measure the correctness of the geometric transformation applied to the object. We leave the design of a better metric for generative object compositing as future work.

\begin{table*}[t]
  \centering
  \small
  \begin{tabular}{lllllll}
    \toprule
    Method & Crop & FID $\downarrow$ & LPIPS $\downarrow$ & CLIP text score $\uparrow$ & CLIP image score $\uparrow$ \\
    \midrule
    BLIP   & \xmark & 18.3673 & 0.0923 & 29.6719 & 95.5625 \\
    SDEdit & \xmark & 17.4963 & 0.0870 & 29.6563 & 96.1250 \\
    \textbf{Ours} & \xmark & \textbf{15.8191} & \textbf{0.0835} & \textbf{29.8594} & \textbf{97.0000} \\
    \midrule
    BLIP   & \cmark & 28.0690 & 0.2463 & 29.0313 & 91.1250 \\
    SDEdit & \cmark & 27.0630 & 0.2312 & 29.0625 & 91.8750 \\
    \textbf{Ours} & \cmark & \textbf{24.4719} & \textbf{0.2223} & \textbf{29.4844} & \textbf{93.7500} \\
    \bottomrule
  \end{tabular}
  \vspace{-5pt}
  \caption{Quantitative comparison with baselines. We measure the performance of our model against two baselines (BLIP and SDEdit) through FID, LPIPS, and modified CLIP scores. The results further demonstrate the effectiveness of our model in addition to the user study results in the paper. More visual comparisons with baselines on the real dataset are shown in \cref{fig:qual0,fig:qual1}.}
  \label{tab:quan}
\end{table*}


\section{Real-World Dataset Collection}
\label{sec:labeling}

\begin{figure*}[t]
\centering
\includegraphics[width=1.0\linewidth]{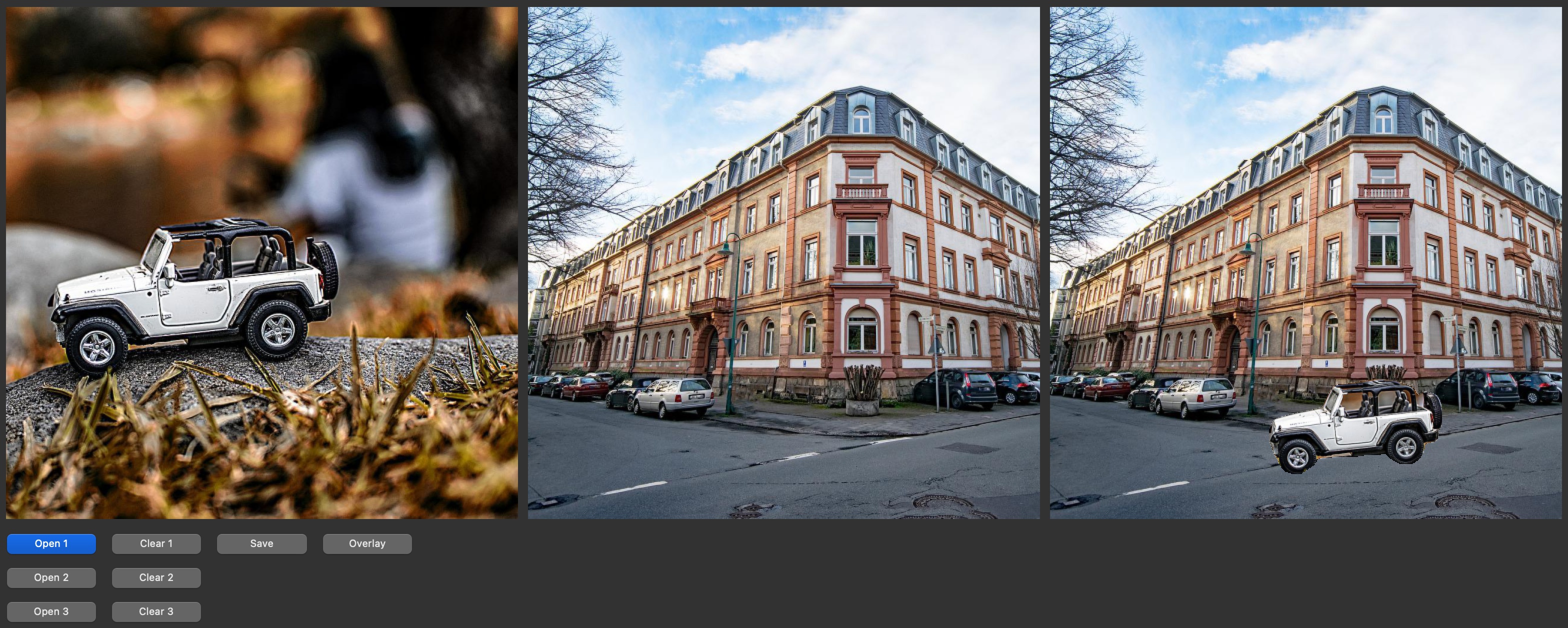} \\
\caption{The user interface of the real-world data labeling tool. This figure demonstrates our data collection process closely simulates the real-world use case in object compositing. The real dataset (including object-background pairs and bounding box annotations) collected using this interface is used for our user study. It consists of three panels: 1) the user can select the object from the left panel; 2) the middle panel shows the background image; 3) the right panel previews copy-and-paste results in real time. Users can drag the object to any location in the background as well as alter the scale of the object.}\label{fig:labeling}
\end{figure*}

\cref{fig:labeling} depicts the real dataset labeling process. The bounding box annotations obtained by this tool are used when generating the result images for our user study. We show that this annotation process directly corresponds to the real-world use case where the user edits a pair of images (an object and a background image) for object compositing. Using this interface, the user can first choose an object image and a background image (displayed in the left panel and the middle panel). Afterward, the user drags the object to a target location on the background image. Then, the location is determined and the object can also be scaled at will. During the whole process, the right panel will display the copy-and-paste image as a preview. Finally, we extract the bounding box of the object to record the location and scale.


\section{Robustness against Low-quality Images}

\begin{figure*}
  \centering
  \begin{subfigure}{0.93\linewidth}
  \includegraphics[width=\textwidth]{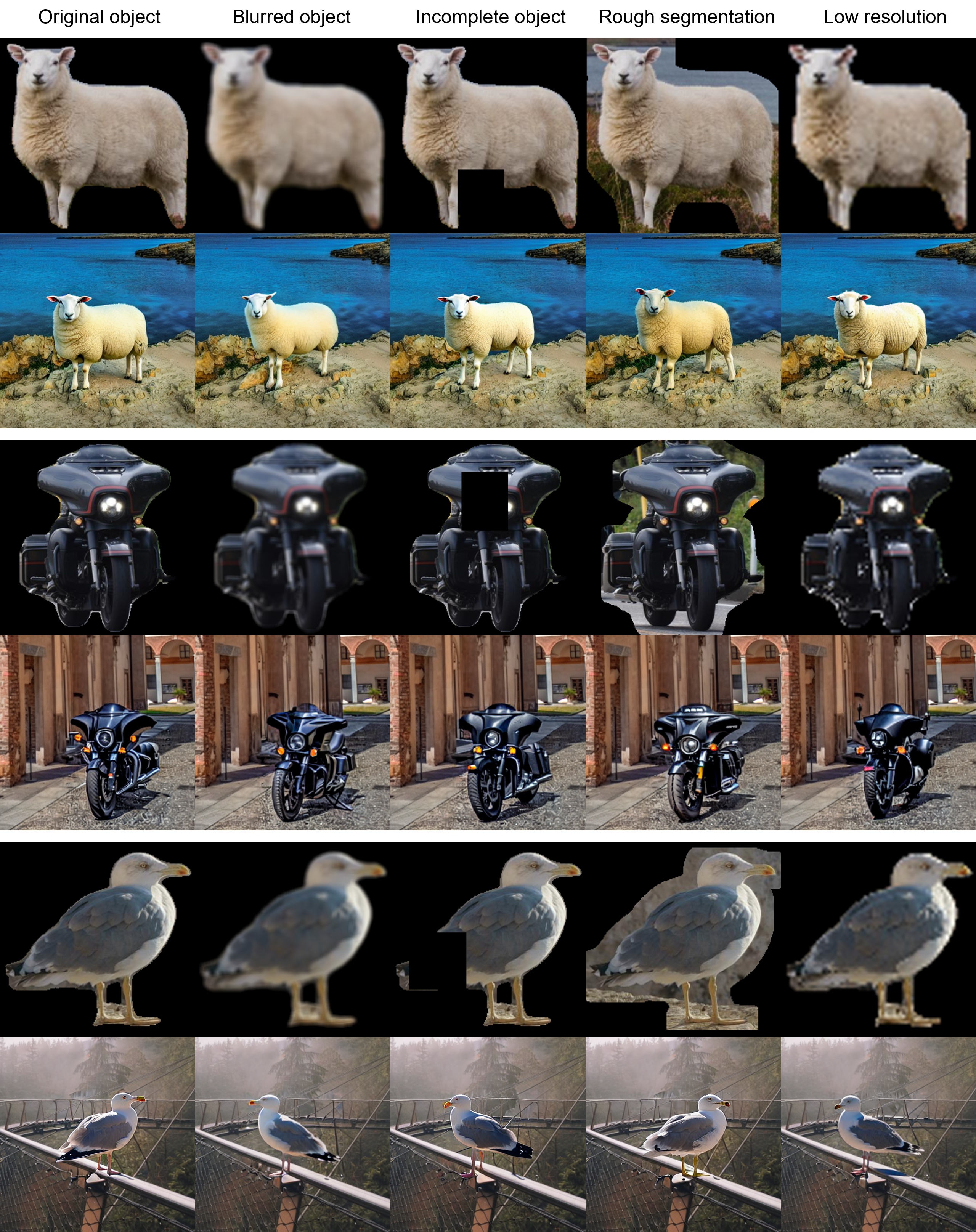}
  \end{subfigure}
  \caption{Robustness. We show our model's robustness against low-quality input objects in the real world. The figure includes three examples. In each example, the top row shows the input object under different conditions: 1) blur, 2) partial occlusion, 3) inaccurate segmentation, and 4) low resolution. The bottom row shows the compositing results corresponding to the input above. Compared to the original input object, our model produces similar high-quality generation results under all conditions.}
  \label{fig:robust1}
\end{figure*}

\begin{figure*}
  \centering
  \begin{subfigure}{0.93\linewidth}
  \includegraphics[width=\textwidth]{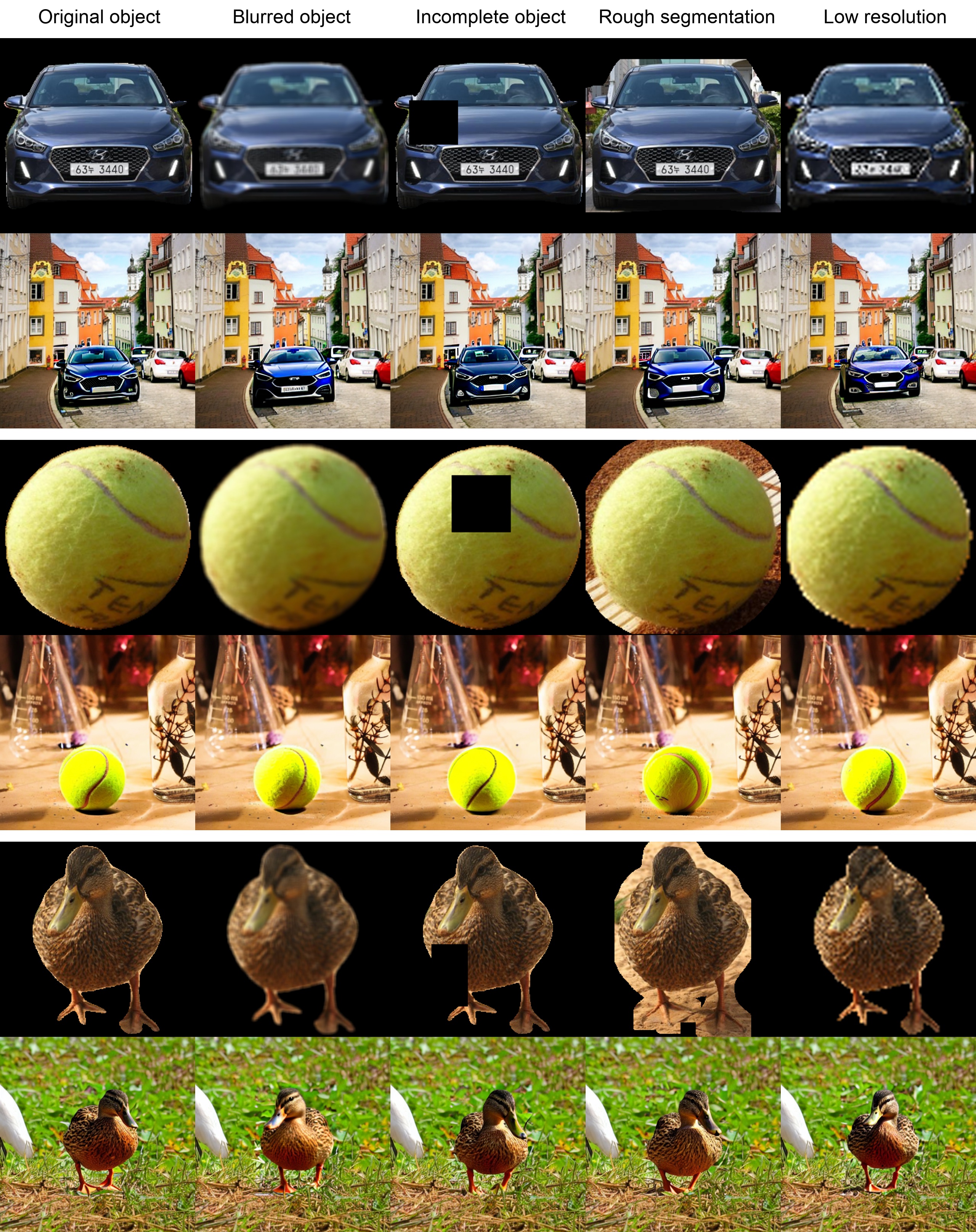}
  \end{subfigure}
  \caption{Robustness. We show our model's robustness against low-quality input objects in the real world. The figure includes three examples. In each example, the top row shows the input object under different conditions: 1) blur, 2) partial occlusion, 3) inaccurate segmentation, and 4) low resolution. The bottom row shows the compositing results corresponding to the input above. Compared to the original input object, our model produces similar high-quality generation results under all conditions.}
  \label{fig:robust2}
\end{figure*}

We mention in the paper that the traditional compositing pipeline \cite{tsai2017deep, sheng2022controllable} cannot address the problem of geometry harmonization and view synthesis, which are advantages of our generation-based method. Another advantage of our model over the traditional pipeline is that it is robust against low-quality input object images. In the real-world scenario of object compositing, it is a common case that the quality of the input object is not perfect. We categorize low-quality input object images into four scenarios:

\begin{itemize}
    \item the input object image is blurry due to lens blur;
    \item some parts of the input object are invisible such as when the object is partially occluded;
    \item the segmentation model fails to extract an accurate segmentation mask of the object, thus the object image includes some background image content; and
    \item the object is too small and thus has low resolution.
\end{itemize}

In our self-supervised training scheme, the synthetic training data we collected covers all the above situations so that the content adaptor will not be constrained by the flaws of the low-level features. In \cref{fig:robust1,fig:robust2} we show examples where low-quality input objects are provided to simulate the aforementioned four scenarios. It is illustrated in the figures that despite the flaws in the input objects, our model is robust to such extreme cases and is still able to generate realistic content.


\section{Additional Qualitative Examples}

\begin{figure*}
  \centering
  \begin{subfigure}{\linewidth}
  \includegraphics[width=\textwidth]{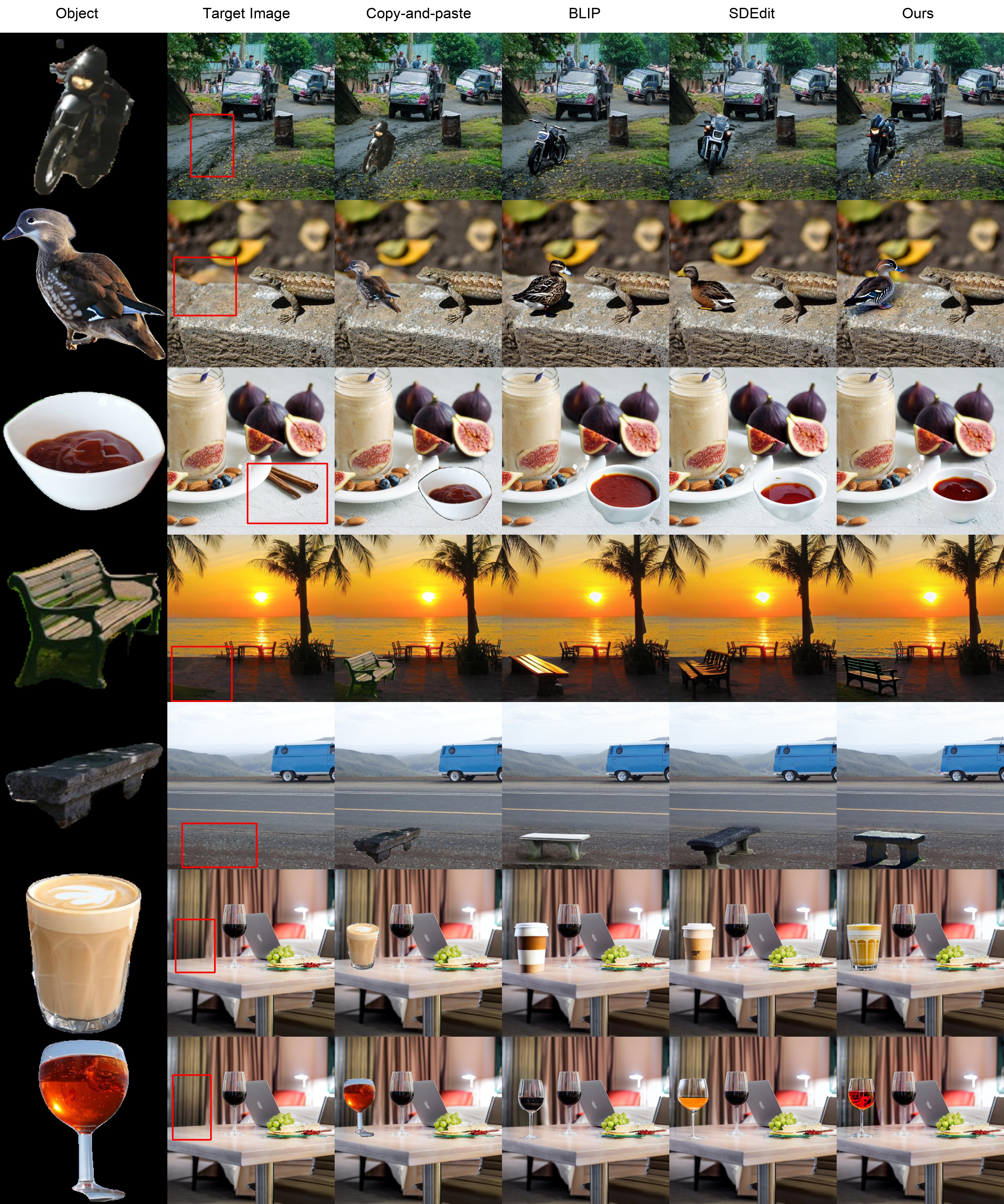}
  \end{subfigure}
  \caption{Qualitative comparison with baseline methods on the real-world test dataset. Our model better \textit{preserves a similar appearance} to the reference object (the first column) while generating \textit{realistic} content that is more consistent with the background.}
  \label{fig:qual0}
\end{figure*}

\begin{figure*}
  \centering
  \begin{subfigure}{\linewidth}
  \includegraphics[width=\textwidth]{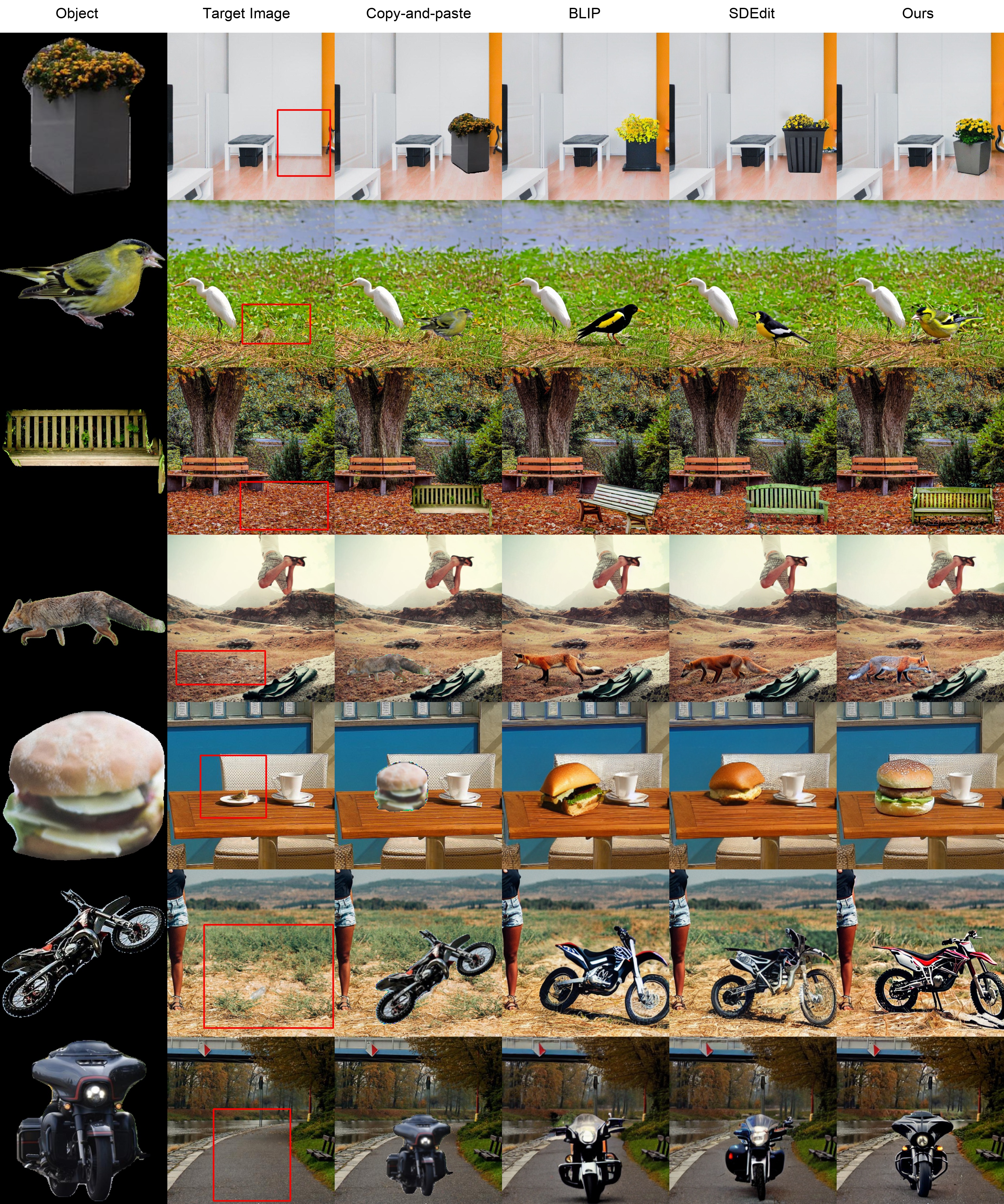}
  \end{subfigure}
  \caption{Qualitative comparison with baseline methods on the real-world test dataset. Our model better \textit{preserves a similar appearance} to the reference object (the first column) while generating \textit{realistic} content that is more consistent with the background.}
  \label{fig:qual1}
\end{figure*}

In addition to the qualitative results in the paper, we show more visual comparisons with baselines in \cref{fig:qual0,fig:qual1}.

\section{Architecture Details and Hyperparameters}

\noindent \textbf{Content Adaptor.}
In the pretraining of the content adaptor, we use a pretrained ViT-L/14 image encoder from \cite{radford2021learning}. This image encoder has been trimmed and only 6 of its 12 attention blocks are kept. We apply this change to better preserve the details of the input object.
In the adaptor, we use one attention layer with 8 heads. Its embedding dimension is 768.

\noindent \textbf{Diffusion steps.}
We use $t=100$ time steps when generating images for the user study and for all qualitative results; $t=50$ is used when testing our model and baselines on the synthetic test data mentioned in \cref{sec:quan}.

\end{appendices}



\end{document}